\documentclass{article}
\usepackage{imav}
\usepackage{times}
\usepackage{graphicx}
\usepackage{amsmath} % assumes amsmath package installed
\usepackage{amssymb} % assumes amsmath package installed
\usepackage{xcolor}
\usepackage{listings}
\usepackage{url}

\definecolor{mGreen}{rgb}{0,0.6,0}
\definecolor{mGray}{rgb}{0.5,0.5,0.5}
\definecolor{mPurple}{rgb}{0.58,0,0.82}
\definecolor{backgroundColour}{rgb}{0.95,0.95,0.92}

\lstdefinestyle{CStyle}{
    backgroundcolor=\color{backgroundColour},   
    commentstyle=\color{mGreen},
    keywordstyle=\color{magenta},
    numberstyle=\tiny\color{mGray},
    stringstyle=\color{mPurple},
    basicstyle=\footnotesize,
    breakatwhitespace=false,         
    breaklines=true,                 
    captionpos=b,                    
    keepspaces=true,                 
    numbers=left,                    
    numbersep=5pt,                  
    showspaces=false,                
    showstringspaces=false,
    showtabs=false,                  
    tabsize=2,
    language=C
}

\title{Guiding vector fields in \emph{Paparazzi} autopilot}
\author{Hector Garcia de Marina\thanks{Email addresseses: hgarciad@ucm.es}, Murat Bronz, and Gautier Hattenberger\thanks{Email addresses: \{murat.bronz, gautier.hattenberger\}@enac.fr} \\ Universidad Complutense de Madrid, Madrid, Spain \\ École National de l'Aviation Civile, Toulouse, France}

\begin{document}

\maketitle
\thispagestyle{empty} % Keep this to remove page number from first page

\begin{abstract}
This article is a technical report on the two different guidance systems based on vector fields that can be found in \emph{Paparazzi}, a free \emph{sw/hw} autopilot. Guiding vector fields allow autonomous vehicles to track paths described by the user mathematically. In particular, we allow two descriptions of the path with an implicit or a parametric function. Each description is associated with its corresponding guiding vector field algorithm. The implementations of the two algorithms are light enough to be run in a modern microcontroller. We will cover the basic theory on how they work, how a user can implement its own paths in \emph{Paparazzi}, how to exploit them to coordinate multiple vehicles, and we finish with some experimental results. Although the presented implementation is focused on fixed-wing aircraft, the guidance is also applicable to other kinds of aerial vehicles such as rotorcraft.
\end{abstract}

\section{Introduction} 
\label{sec: intro}
Autonomous aerial vehicles are presented as a great assistance to humans in challenging tasks such as environmental monitoring, search \& rescue, surveillance, and inspection missions \cite{yang2018grand}. As mobile vehicles, they are \emph{typically} commanded to travel from \emph{point A} to \emph{point B}. Even more, the requirements of the task at hand might demand a more precise route or \emph{path} to be tracked while traveling from \emph{A} to \emph{B}. Guiding vector fields allow autonomous vehicles to track desired paths accurately with any temporal restriction. For example, we only assign the vehicle to visit a collection of connected points in the space (a geometric object); thus, we do not concern ourselves when the vehicle visits a specific point of such a geometric object. Guiding vector fields have been widely studied and employed in many different kinds of vehicles \cite{Goncalves2010,rezende2018robust,michalek2018vfo,KAPITANYUK20147342,lakomy2017,michalek2010vector}.

Two guiding vector fields have been implemented in \emph{Paparazzi}, an open-source drone hardware and software project encompassing autopilot systems and ground station software for multicopters/multirotors, fixed-wing, helicopters and hybrid aircraft that was founded in 2003 \cite{hattenberger2014using}.

The first guiding vector field, or simply \emph{GVF}, allows fixed-wing aircraft to track 2D (constant altitude) paths described by an implicit equation in \emph{Paparazzi} \cite{de2017guidance}. This \emph{GVF} guidance system has been exploited to coordinate a fleet of aircraft on circular paths \cite{de2017circular}. The second guidance system is the parametric guiding vector field, or simply \emph{p-GVF} \cite{yao2021singularity}. This evolved version allows fixed-wing aircraft to track 3D paths described by a parametric equation in \emph{Paparazzi}, and it also allows the coordination of multiple vehicles \cite{yao2021distributed}. The main feature of the \emph{p-GVF} is that it allows for tracking paths that are self-intersected, such an \emph{eight figure}, and guarantees global convergence to the path. This \emph{p-GVF} guidance systems has been exploited to characterize soaring planes.

Both implementations compensate for the disturbance of the wind on the vehicle by crabbing. Crabbing happens when the inertial velocity makes an angle with the nose heading due to wind. \emph{Slipping} occurs when the aerodynamic velocity vector makes an angle (sideslip) with the body ZX plane. Slipping is (almost) always undesirable because it degrades aerodynamic performance. Crabbing is not an issue for the aircraft.

We split this paper into two equal parts focused on each guidance system. We will briefly show how the \emph{GVF} and the \emph{p-GVF} work and how they are implemented in \emph{Paparazzi} so that a final user can define and try his own trajectories for fixed-wing aircraft. We present some performance results from actual telemetry, and we end with demonstrations concerning the coordination of more than one aircraft.

\section{The \emph{GVF} guidance system}
\subsection{Theory}
\begin{figure}
\centering
\includegraphics[width=1\columnwidth]{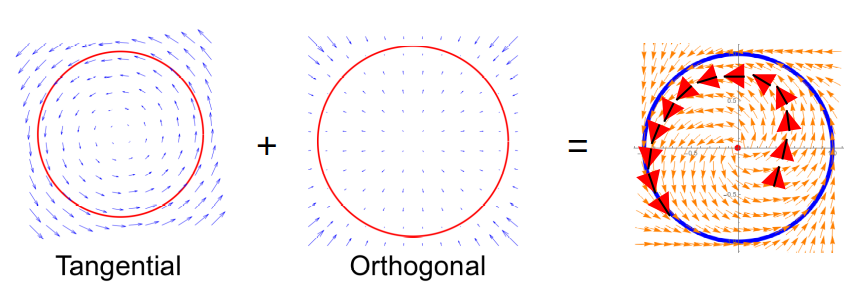}
	\caption{The \emph{GVF} combines the vectors tangential and orthogonal to the level set of the desired path. The orthogonal part always points towards the desired path if the gradient of the level set is multiplied by the error quantity (\emph{current level set - desired level set}).}
\label{fig: gvf}
\end{figure}

This guidance system is based on constructing vectors tangential to the different level sets of the path in implicit form. Then, we add a normal component facing towards the direction of the desired level set. This will make a guiding vector that will drive the vehicle smoothly to travel on the desired path. Since we only care about the direction to follow and not the speed, we normalize the result to track a unit vector. This rationale is explained in Figure \ref{fig: gvf}. We can express this technique formally as follows
\begin{equation}
	\dot p_d(p) := \tau(p) - k_e e(p) n(p),
	\label{eq: gvf}
\end{equation}
where $\frac{\dot p_d}{||\dot p_d||} \in \mathbb{R}^2$ is the unit vector to follow, $e\in\mathbb{R}$ is the current level set, $\tau\in\mathbb{R}^2$ and $n\in\mathbb{R}^2$ are the tangent and normal vectors respectively to the current level set, and $k_e > 0$ is a positive gain that defines how aggresive is the convergence of the guiding vector to the desired path. Note that all the variables in (\ref{eq: gvf}) depend on the position $p\in\mathbb{R}^2$ of the aircraft.

In order to align the vehicle's velocity with the vector field in (\ref{eq: gvf}), the control action in \emph{Paparazzi} will need of the gradient and the Hessian of the desired path \cite{de2017guidance}. In \emph{Paparazzi} there is another gain $k_n$ for a proportional controller to align both, the current velocity and the desired velocity.

\subsection{Paths for \emph{GVF} in Paparazzi}
Let us illustrate this section with an example, and we will use it to demonstrate how the user has to write code to implement an arbitrary path in \emph{Paparazzi}. Let us focus on a circle whose implicit equation is given by
\begin{equation}
	\mathcal{P}(x,y) := x^2 + y^2 - r^2 = 0,
\end{equation}
where $r$ is the radius of the circle, and $x$ and $y$ the standard Cartesian coordinates. We first note that we define the desired level set as the zero level set. Therefore, the variable $e$ in (\ref{eq: gvf}) can be identified as $e = \mathcal{P}(x,y)$. The gradient or $n(p)$ in (\ref{eq: gvf}) is trivially calculated as $n(p) = \begin{bmatrix}2x & 2y\end{bmatrix}$, and the Hessian $H(p) = \begin{bmatrix}2 & 0 \\ 0 & 2 \end{bmatrix}$.

All these path-dependent quantities must be codified in a file called \emph{gvf/trajectories/gvf\_circle.c}\footnote{The code can be found at \url{https://github.com/paparazzi/paparazzi/tree/master/sw/airborne/modules/guidance}}.

\begin{lstlisting}[style=CStyle]
void gvf_circle_info(float *phi, struct gvf_grad *grad, struct gvf_Hess *hess)
{

  struct EnuCoor_f *p = stateGetPositionEnu_f();
  float px = p->x;
  float py = p->y;

  // Parameters of the trajectory, circle's center and radius
  float wx = gvf_trajectory.p[0];
  float wy = gvf_trajectory.p[1];
  float r = gvf_trajectory.p[2];

  // Phi(x,y) or signal e
  *phi = xrel*xrel + yrel*yrel - r*r;

  // grad Phi
  grad->nx = 2 * xel;
  grad->ny = 2 * yel;

  // Hessian Phi
  hess->H11 = 2;
  hess->H12 = 0;
  hess->H21 = 0;
  hess->H22 = 2;
}
\end{lstlisting}

Then, the user needs to define a high-level function in \emph{gvf/gvf.c} to be called from the flight plan as follows

\begin{lstlisting}[style=CStyle]
bool gvf_circle_XY(float x, float y, float r)
{
  float e;
  struct gvf_grad grad_circle;
  struct gvf_Hess Hess_circle;

  gvf_trajectory.type = 1; // It is a circle
  gvf_trajectory.p[0] = x;
  gvf_trajectory.p[1] = y;
  gvf_trajectory.p[2] = r;

  gvf_circle_info(&e, &grad_circle, &Hess_circle);
  gvf_control.ke = gvf_circle_par.ke;
  gvf_control_2D(gvf_circle_par.ke, gvf_circle_par.kn, e, &grad_circle, &Hess_circle);

  gvf_control.error = e; // For telemetry

  return true;
}
\end{lstlisting}

The function \emph{gvf\_control\_2D} calculates the desired roll angle to be tracked by the aircraft in order to align its velocity to (\ref{eq: gvf}). With only the definition of these two functions, together with the corresponding definitions in headers for the gains and used structs, is how a new path for the \emph{GVF} guidance system is defined in \emph{Paparazzi}.

\subsection{Performance in Paparazzi}
The figure \ref{fig: gcsgvf} shows the described trajectory of an aircraft tracking a 2D ellipse in a windy environment. The airspeed of the aircraft was around 11m/s, and the windspeed was around 5 m/s.

\begin{figure}
\centering
\includegraphics[width=0.7\columnwidth]{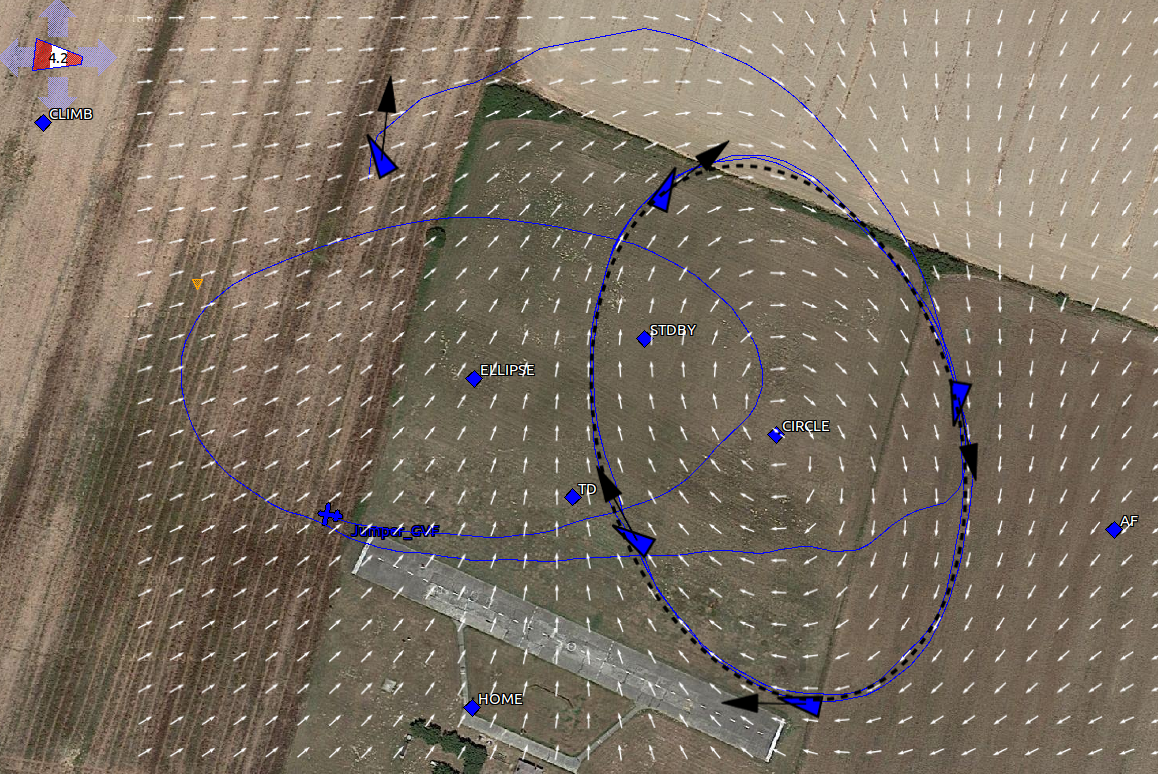}
	\caption{The \emph{Paparazzi} ground control station showing different times for the position of an aircraft tracking a 2D ellipse with the \emph{GVF} guidance system. The white arrows construct the vector field with unitary vectors calculated from (\ref{eq: gvf}).}
\label{fig: gcsgvf}
\end{figure}

\subsection{Multi-vehicles}
When the desired path is closed, such as a circle, then we can exploit the convergence properties of the \emph{GVF} to synchronize different aircraft on the path. The different aircraft have to follow a positive or negative level set with respect to the desired path. While following a negative/positive level set, then the vehicle travels inside/outside the desired path and travels a shorter/larger distance in one lap. In that way, the aircraft can catch up or separate from each other. This can be done in a distributed way without any intervention from the Ground Control station. This implementation in \emph{Paparazzi} has been discussed in detail in \cite{de2017distributed} and the theory can be found in \cite{de2017circular}.
\begin{figure}
\centering
\includegraphics[width=0.7\columnwidth]{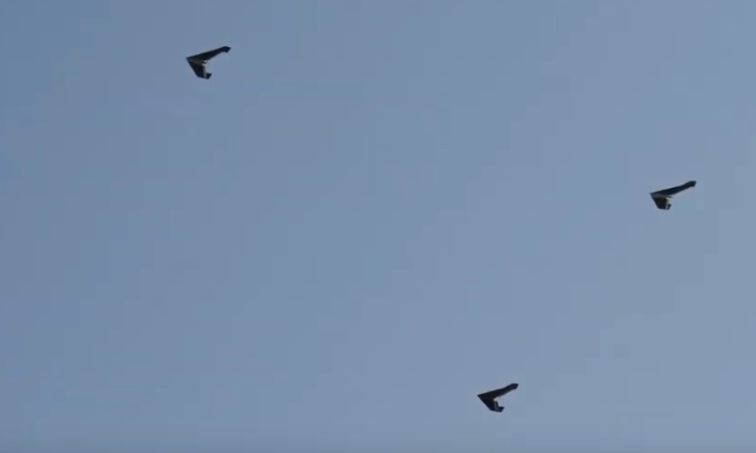}
	\caption{The \emph{GVF} in \emph{Paparazzi} can be employed to synchronize multiple aircraft in a distributed way. This caption corresponds to the 2017 IMAV competition.}
\label{fig: gcsgvf}
\end{figure}

\section{The \emph{p-GVF} guidance system}
\subsection{Theory}
\begin{figure}
\centering
\includegraphics[width=1\columnwidth]{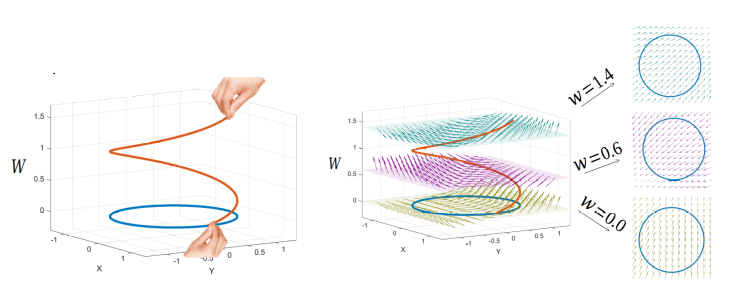}
	\caption{The p-GVF solves the singularity problem by taking a parametric description of the desired path. Then, the parameter $w$ becomes a virtual dimension (topological surgery on the left image) so that we construct a singularity-free vector field following the tangential+orthogonal vector approach. The vehicle only needs to follow the projection of the vector field on the \emph{physical world} coordinates.}
\label{fig: pgvf}
\end{figure}

The guidance system \emph{GVF} has a big inconvenience; the vector field (\ref{eq: gvf}) is not defined when the gradient is zero. We call this situation a singularity. For example, if $x = y = 0$ for the circle. That makes it impossible to implement paths with self-intersections. To avoid this difficulty, we have developed the \emph{parametric Guiding Vector Field} or \emph{p-GVF} \cite{yao2021singularity}. We start from the parametric equation of the desired path, for example, for the circle
\begin{equation}
	\begin{cases}
	x &= r\cos w \\
    y &= r\sin w
	\end{cases},
\end{equation}
where $w\in\mathbb{R}$ is a free parameter. Then, we consider a $(2+1)$ dimensional path (two physical dimensions, and one virtual for $w$) as on the left side in Figure \ref{fig: pgvf}. Now, this new higher dimensional path can be seen in its implicit form for each coordinate so that we can apply a similar technique as in (\ref{eq: gvf}) again. In particular, the new implicit form for the circle is
\begin{equation}
e_x = x - r\cos w, \quad e_y = y - r\sin w,
\end{equation}
and the vector field to be followed has the form
\begin{equation}
	\xi = \nabla_\times e - \sum_{i=\{x,y\}} k_i e_i \nabla e_i,
\end{equation}
where the first term to be explained shortly makes the vehicle to follow the path (\emph{tangential component}), and the second one makes the vehicle to approach the path (\emph{normal component}). The variable $e = \begin{bmatrix}e_x & e_y\end{bmatrix}^T$, $\nabla$ is the gradient operator (note that $\nabla e_i = \begin{bmatrix}0,\dots,1,\dots,\frac{\partial e_i}{\partial w}\end{bmatrix}$), and 
	\begin{equation}
		\nabla_\times e = (-1)^n \begin{bmatrix}\frac{\partial e_1}{\partial w} & \frac{\partial e_2}{\partial w} & \dots & \frac{\partial w}{\partial w}\end{bmatrix}^T.
	\end{equation}
Note that the last term $\frac{\partial w}{\partial w}$ sets the eventual velocity for $w$ to one. Eventually, in \emph{Paparazzi}, the desired velocity for $w$ adapts to the actual speed of the vehicle. The main difference with respect to \emph{GVF} is that the resultant guiding vector is not only driving the Cartesian coordinates but the virtual coordinate $w$. This can be seen on the right-hand side in Figure \ref{fig: pgvf}. We are free of singularities with this technique.

\subsection{Paths for \emph{p-GVF} in Paparazzi}
Similarly, a user will need to define two main functions for an arbitrary path. The first one defines the parametric/implicit equations of the trajectory, i.e., $e_x, e_y,$ and $e_z$, and its partial derivatives with respect to $w$ for a 3D path. In the following example, we take a tilted circle where $z_h$ and $z_l$ define the maximum and minimum altitude of the circle of radius $r$. This function would be placed at \emph{gvf\_parametric/trajectories/gvf\_parametric\_3d\_ellipse.c}.

\begin{lstlisting}[style=CStyle]
void gvf_parametric_3d_ellipse_info(float *f1, float *f2, float *f3, float *f1d, float *f2d, float *f3d, float *f1dd, float *f2dd, float *f3dd)
{
  float xo = gvf_parametric_trajectory.p_parametric[0];
  float yo = gvf_parametric_trajectory.p_parametric[1];
  float r = gvf_parametric_trajectory.p_parametric[2];
  float zl = gvf_parametric_trajectory.p_parametric[3];
  float zh = gvf_parametric_trajectory.p_parametric[4];
  float alpha_rad = gvf_parametric_trajectory.p_parametric[5]*M_PI/180;

  float w = gvf_parametric_control.w;
  float wb = w * gvf_parametric_control.beta * gvf_parametric_control.s;

  // Parametric equations of the trajectory and the partial derivatives w.r.t. 'w'

  // These are e_x, e_y, and e_z
  *f1 = r * cosf(wb) + xo;
  *f2 = r * sinf(wb) + yo;
  *f3 = 0.5 * (zh + zl + (zl - zh) * sinf(alpha_rad - wb));

  // These are the partials of e_x, e_y, and e_z w.r.t. 'w'
  *f1d = -r * sinf(wb);
  *f2d = r * cosf(wb);
  *f3d = -0.5 * (zl - zh) * cosf(alpha_rad - wb);

  // These are the second partials of e_x, e_y, and e_z w.r.t. 'w'
  *f1dd = -r * cosf(wb);
  *f2dd = -r * sinf(wb);
  *f3dd = -0.5 * (zl - zh) * sinf(alpha_rad - wb);
}
\end{lstlisting}
Secondly, the following high-level function to be called from the flight plan must be placed at \emph{guidance/gvf\_parametric/gvf\_parametric.cpp}.

\begin{lstlisting}[style=CStyle]
bool gvf_parametric_3D_ellipse_XYZ(float xo, float yo, float r, float zl, float zh, float alpha)
{
  gvf_parametric_trajectory.type = ELLIPSE_3D;
  gvf_parametric_trajectory.p_parametric[0] = xo;
  gvf_parametric_trajectory.p_parametric[1] = yo;
  gvf_parametric_trajectory.p_parametric[2] = r;
  gvf_parametric_trajectory.p_parametric[3] = zl;
  gvf_parametric_trajectory.p_parametric[4] = zh;
  gvf_parametric_trajectory.p_parametric[5] = alpha;

  float f1, f2, f3, f1d, f2d, f3d, f1dd, f2dd, f3dd;

  gvf_parametric_3d_ellipse_info(&f1, &f2, &f3, &f1d, &f2d, &f3d, &f1dd, &f2dd, &f3dd);
  gvf_parametric_control_3D(gvf_parametric_3d_ellipse_par.kx, gvf_parametric_3d_ellipse_par.ky, gvf_parametric_3d_ellipse_par.kz, f1, f2, f3, f1d, f2d, f3d, f1dd, f2dd, f3dd);

  return true;
}
\end{lstlisting}
Note that we have a collection of positive gains to be tuned: $k_x, k_y,$ and $k_z$, for the convergence of the different Cartesian coordinates.

\subsection{Performance in Paparazzi}
\begin{figure}
\centering
\includegraphics[width=0.45\columnwidth]{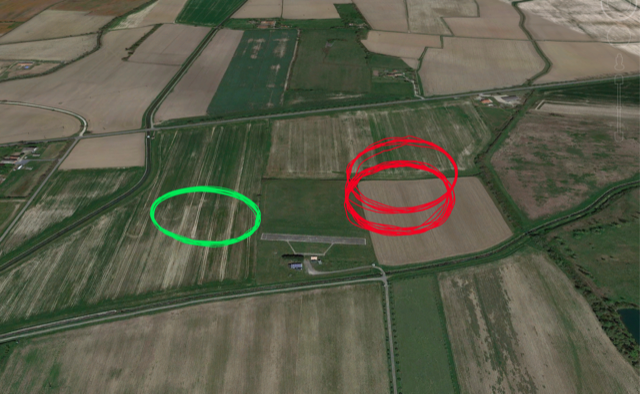}
\includegraphics[width=0.45\columnwidth]{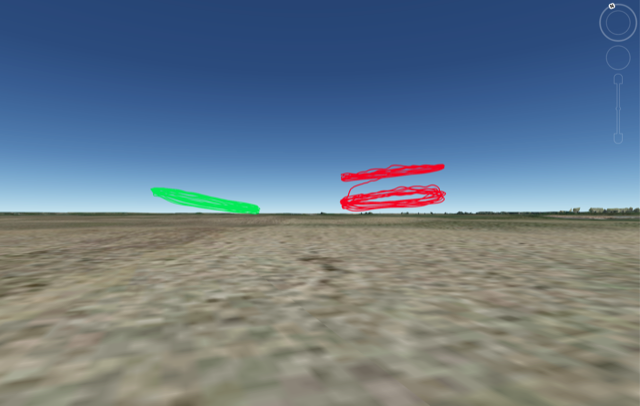}
	\caption{The \emph{g-GVF} guidance system in \emph{Paparazzi} allows aircraft to track 3D paths. These flights correspond to dynamic soaring experiments where the aircraft was tracking a tilted circle.}
\label{fig: pgvfgcs}
\end{figure}

In figure \ref{fig: pgvfgcs} we show the performance of one aircraft tracking a simple Lissajous figure that is bent in 3D. The \emph{p-GVF} controller passes two setpoints to low-level controllers, namely, the desired heading rate and the desired vertical speed. The first one is handled by controlling the roll angle of the aircraft depending on its actual ground speed. The second one is handled by controlling both the pitch and the throttle of the aircraft. Therefore, in order to have a good performance in \emph{Paparazzi}, the vertical speed controller must be tuned in advance.

\subsection{Multi-vehicles}
\begin{figure}
\centering
\includegraphics[width=1\columnwidth]{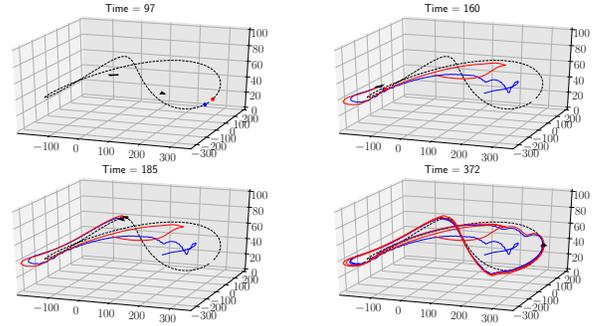}
	\caption{The \emph{g-GVF} guidance system in \emph{Paparazzi} allows aircraft to synchronize aircraft on 3D paths. In this example, both aircraft are instructed to have the same $w$ so that they rendezvous.}
\label{fig: pgvfm}
\end{figure}

We can employ the \emph{p-GVF} guidance system to synchronize vehicles on the desired path. Differently than with the \emph{GVF}, now we will focus on controlling the \emph{distances} between the virtual coordinates $w$ for each vehicle. This is done by injecting the standard consensus algorithm to the control action, where each aircraft share their current $w$ and compares the subtraction with the desired value \cite{yao2021distributed}. The result makes the vehicles travel on the desired \emph{parametric} path with their desired relative $w$ between each other. This algorithm can run a distributed way, and there is no need for a Ground Control station in \emph{Paparazzi}. In figure \ref{fig: pgvfm} we show the rendezvous of two aircraft on the same 3D path.

\section{Conclusions and future work}
This article presents a technical report on the two guiding systems based on guiding vector fields available in \emph{Paparazzi} autopilot. We have shown how the user can implement its own desired path by mainly creating two functions in C code. The first function contains the mathematical information of the desired path, while the second function is what is called by the user from the Ground Control station. The presented guiding systems can be employed in \emph{Paparazzi} to coordinate aircraft in a distributed way.

The future work focuses on extending the implementation in \emph{Paparazzi} to other vehicles such as rotorcraft and rovers.

\section*{Acknowledgements}
The work of Hector Garcia de Marina is supported by the grant \emph{Atraccion de Talento} with reference number 2019-T2/TIC-13503 from the Government of the Autonomous Community of Madrid.

% BIBLIOGRAPHY:
% use {unsrt}:
\bibliographystyle{unsrt}
\bibliography{imav_bibliography}

% The following lines are necessary for showing the appendices correctly, do not change!
\appendix
\newcommand{\appsection}[1]{\let\oldthesection\thesection
  \renewcommand{\thesection}{Appendix \oldthesection:}
  \section{#1}\let\thesection\oldthesection}
% appendices are now indicated by appsection:

\end{document}